\documentclass{article}

\usepackage{PRIMEarxiv}
\usepackage{authblk}
\usepackage[utf8]{inputenc} % allow utf-8 input
\usepackage[T1]{fontenc}    % use 8-bit T1 fonts
\usepackage{hyperref}       % hyperlinks
\usepackage{url}            % simple URL typesetting
\usepackage{booktabs}       % professional-quality tables
\usepackage{amsfonts}       % blackboard math symbols
\usepackage{nicefrac}       % compact symbols for 1/2, etc.
\usepackage{microtype}      % microtypography
\usepackage{lipsum}
\usepackage{fancyhdr}       % header
\usepackage{graphicx}       % graphics
\graphicspath{{media/}}     % organize your images and other figures under media/ folder
\usepackage{graphicx}
\usepackage{subfigure}
\usepackage{todonotes}
\usepackage{cleveref}
\usepackage{multirow}
\usepackage{amssymb}% http://ctan.org/pkg/amssymb
\usepackage{pifont}% http://ctan.org/pkg/pifont
\usepackage[makeroom]{cancel}
\usepackage{xcolor}
\usepackage[flushleft]{threeparttable}
\newcommand{\cmark}{{\color{green} \ding{51}}}%
\newcommand{\xmark}{{\color{red}\ding{55}}}%
\newcommand{\pmark}{{\color{blue}\bcancel{\ding{51}}}}%

\title{The Framework That Survives Bad Models:\\Human-AI Collaboration For Clinical Trials}

\author{
  Yao Chen, David Ohlssen, Aimee Readie, Gregory Ligozio, Ruvie Martin,  Thibaud Coroller\textsuperscript{@} \\
  Novartis Pharmaceuticals Corporation, NJ, USA\\
  \textsuperscript{@}Corresponding author: thibaud.coroller@novartis.com \\
}

\begin{document}

\maketitle

\begin{abstract}
% Artificial intelligence (AI) is a promising tool for clinical trials. In this study, we explore the use of AI frameworks to classify spinal image for modified Stoke Ankylosing Spondylitis Spinal Score (mSASSS). We also investigate the usage of poorly performing model and the safety of our framework to mitigate it. We found that AI as supporting reader (AI-SR) was the most suitable approach for clinical trial as it satisfied all criteria, under all model types \ref{tab:summary}. This approach also had accurate measurement for the disease estimation and the clinical trial treatment effect.

Artificial intelligence (AI) holds great promise for supporting clinical trials, from patient recruitment and endpoint assessment to treatment response prediction. However, deploying AI without safeguards poses significant risks, particularly when evaluating patient endpoints that directly impact trial conclusions. We compared two AI frameworks against human-only assessment for medical image-based disease evaluation, measuring cost, accuracy, robustness, and generalization ability. To stress-test these frameworks, we injected bad models—ranging from random guesses to naive predictions—to ensure that observed treatment effects remain valid even under severe model degradation. We evaluated the frameworks using two randomized controlled trials with endpoints derived from spinal X-ray images. Our findings indicate that using AI as a supporting reader (AI-SR) is the most suitable approach for clinical trials, as it meets all criteria across various model types, even with bad models. This method consistently provides reliable disease estimation, preserves clinical trial treatment effect estimates and conclusions, and retains these advantages when applied to different populations.
\end{abstract}

% keywords can be removed

\section{Introduction}

Today, AI has become an integrated part of our life, from large language models (LLMs) \cite{touvron2023llama, guo2025deepseek}, car navigation, to accurate weather forecasting. To become an integrated part of our life, those models have shown exceptional performance (i.e., models' ability to successfully complete a task), and the risk of any failure is greatly outweighed by the value they bring. However, in comparison, AI has had little impact on healthcare and drug development despite multiple areas where it can help. First, AI has been suggested for data analysis of vast amounts of data generated by clinical trials \cite{fournier2021incorporating}. It can help researchers identify potential drug candidates, predict patient responses, and optimize trial designs. This can lead to personalized medicine, leading to more effective and targeted therapies. Secondly, AI has the potential to enhance the recruitment of suitable patients \cite{harrer2019artificial}, one of the biggest challenges in clinical trials. AI can help identify and recruit eligible patients more efficiently \cite{zhang2023harnessing} by analyzing electronic health records, omics information, and other relevant non-traditional data sources. And finally, disease assessment \cite{yang2022artificial} using AI to evaluate disease progression (e.g., grading medical images ) can greatly reduce clinical trial costs and time, as well as reducing error (reader variability, human entry error).

To bring AI to clinical trials, researchers can either try to improve model performance to match human performance and place all risk on AI model alone, or mitigate AI risks in clinical trials, accepting the model may perform worse than human readers. In this study we will focus on the latter, understanding how a trained model (as high performance as possible) will perform if used for clinical trials. We also explored the cases where the AI model is dangerous (e.g. random prediction) in order to stress test the limit of our framework and ensure strong safeguards. Inspired by the study from Ng \textit{ et al.}\cite{ng2023artificial} which looked at using AI to classify cancer images (human vs AI-independent vs. AI-supported), we modified the concept to be applied in immunology to classify spinal X-ray images. We further extended their original scope by quantifying the treatment effect (key endpoint in clinical trial). In summary, in this study we aim to quantify several key aspects of using AI in clinical trials, namely:

\begin{enumerate}
    \item \textbf{Efficiency}: Can AI reduce clinical trial time and cost?
    \item \textbf{Robustness}: Is using a moderately performing AI detrimental to the disease evaluation? To what extent is the model's performance acceptable for safe use?
    \item \textbf{Generalization}: Can this framework be applied to more than one study?
    \item \textbf{Accuracy}: Does use of AI lead to the same clinical finding at the clinical trial level (e.g. similar treatment effect estimation, same study conclusion)?
\end{enumerate}

\section{Methods}
\subsection{Data}

In this work, we used X-ray data from two phase III clinical studies, MEASURE I and PREVENT to test the effectiveness and generalization of the frameworks. Our AI grading model described later in \Cref{sec:model} was trained on MEASURE I, predictions from the validation sets of MEASURE I and PREVENT, which guarantee the independence with the model were used for the test of the framework.

\begin{figure}[!htb]
    \centering
    \includegraphics[width=0.6\textwidth]{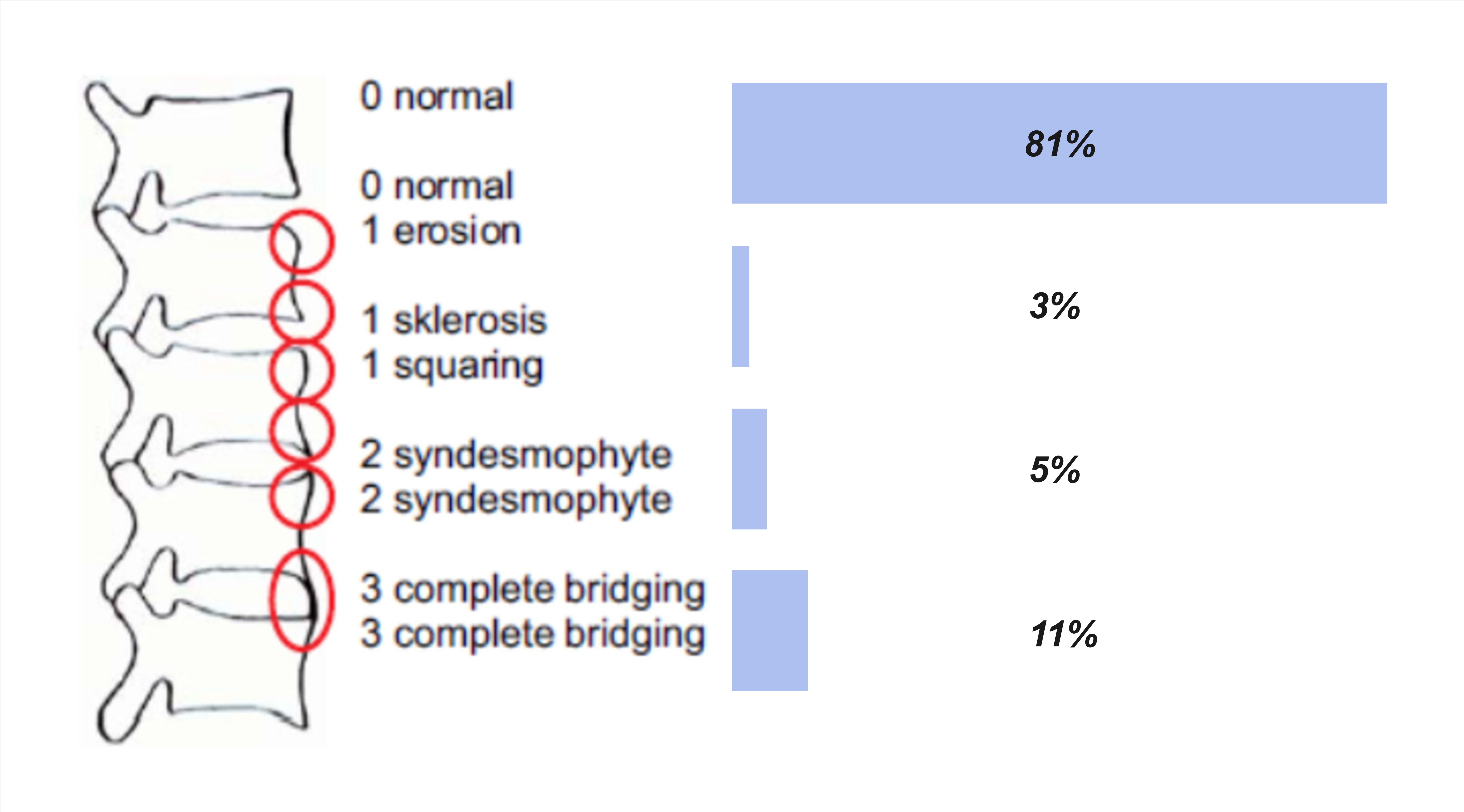}
    \caption{mSASSS definition and distribution in the dataset. mSASSS scoring criteria (0–3) based on structural changes at vertebral corners, alongside their frequency in the dataset. Most corners are normal (score 0, 81\%), while erosion/sclerosis/squaring (score 1), syndesmophytes (score 2), and bridging (score 3) are less common (3\%, 5\%, and 11\%, respectively).}
    \label{fig:mSASSS}
\end{figure}

MEASURE I\cite{Measure1} is a completed and anonymized phase III trial for secukinumab (a fully human anti-IL-17A monoclonal antibody) in patients with axSpA (axial spondyloarthritis). MEASURE I included 361 patients randomized to the treatment and control groups in a 2:1 ratio. Patients on the control arm started with placebo and then switched to active treatment after 16 or 24 weeks, depending on their responder status. Two (or three X-ray) sessions were scheduled for each patient at baseline, 2 years (Week 104) and 4 years (Week 208). Lateral cervical and lumbar X-rays were taken during those sessions. The modified Stoke Ankylosing Spondylitis Spinal Scores (mSASSS) in \Cref{fig:mSASSS} were evaluated by at least two radiologists, and exploratory endpoints were established to assess the worsening of mSASSS from baseline.

\graphicspath{./media/}

% endpoint definition
In this study, the mSASSS worsening after two years (between baseline and week 108 X-rays) was defined as an exploratory endpoint to determine the treatment effect between the treatment and placebo groups. 
% reading scheme
Two reading sessions were conducted by two radiologists. During each session, the mSASSS assessment was conducted at each vertebral corner. The total score was calculated by summing the mSASSS values across 24 corners. 
Two reading sessions were conducted that aligned with the collection of the Week 104 and Week 208 images. During each reading session, the mSASSS was evaluated by two radiologists, each independently scoring the vertebral corners for structural changes. The total score was calculated by summing the mSASSS values across the 24 locations. To assess for inter-rater variability, the top 5\% most discrepant total score cases between the two readers were submitted for adjudication. The distribution of mSASSS in each vertebral corner is shown in \Cref{fig:mSASSS}, indicating a noticeable imbalance in healthy vertebrae versus those with structural changes due to disease progression.
% In the first session, all X-rays were read, and then around 60\% X-rays were randomly selected to be read again in the second session. If a large discrepancy was detected in the total patient mSASSS, an adjudication would be introduced to make the final decision. The distribution of mSASSS in each vertebral corner is shown in \Cref{fig:mSASSS}, showing a great imbalance in healthy vertebrae vs. non-healthy ones. 

% PREVENT study
Similarly to MEASURE I, PREVENT \cite{prevent} is another trial of secukinumab in axSpA but with a different patient population, that of non-radiographic axSpA. PREVENT collected the same spinal X-rays and shared the same reading schemes as MEASURE I. However, PREVENT is very different from MEASURE I in terms of patient population. The population in PREVENT is much less severe. In fact, more than 90\% of the patients have zero mSASSS in PREVENT, which makes it a perfect trial to test the generalization ability of the candidate frameworks.

\subsection{Framework}
\label{sec:framework}

\begin{figure}[h]
    \centering
    \includegraphics[width=0.9\textwidth]{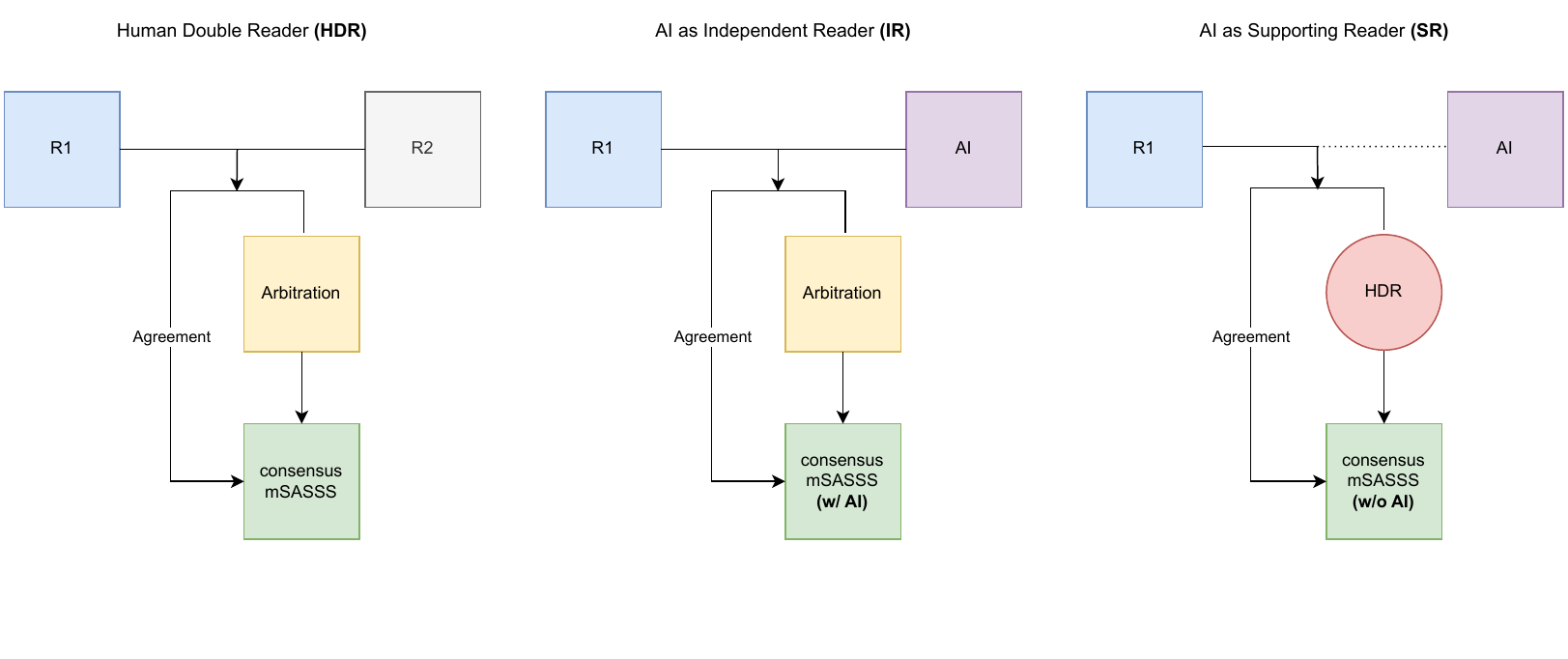}
    \caption{Introducing the framework implementation for the experiment, where we compare the gold standard, human double reader (\textbf{HDR}) to two AI approaches, namely AI as independent (\textbf{IR}) or supporting reader (\textbf{SR}) to evaluate spinal disease status (modified Stoke Ankylosing Spondylitis Spinal Scores - mSASSS) from X-ray images. At the end of each framework, the scores from each reader are pooled to obtain the consensus mSASSS. One key difference between the two AI frameworks, is that AI-IR includes the AI score for the consensus mSASSS, while AI-SR does not.}
    \label{fig:framework}
\end{figure}

Our framework (Figure \ref{fig:framework}) adapted the Ng \textit{et al}\cite{ng2023artificial} framework to grade spinal images. For each patient, two (or more) images were graded and a composite score was computed (mSASSS) using all vertebral corners. To mimic the human process as closely as possible, we only compared total spine mSASSS as opposed to individual vertebra levels. As with human graders, this approach focuses on the consistency of the total score rather than mSASSS at each individual vertebral corner. In such a setting, the Human Double reader in \Cref{fig:framework} is the current clinical application, while the two others (AI-independent and AI-supported) are the experimental methods we are scoping in this work. The AI model utilized in this test is detailed in \Cref{sec:model}. If the AI model fails to predict the mSASSS, this missingness will automatically trigger a disagreement, necessitating the involvement of at least two human readers. Additionally, we explored the use of unreliable AI models within these frameworks to evaluate the robustness of the outcomes in the following sections.

\graphicspath{{./media/}}

% We also investigated the use of non-reliable AI models within this framework to assess the robustness of the outcomes. Several rules are set to mitigate using such models, namely:

% \begin{itemize}
%     \item \textbf{Data missingness:} Missing AI spinal score automatically triggers arbitration, so it always requires at least two human readers.
%     \item \textbf{AI contribution:} AI is not equal to human readers for AI-SR (excluded from consensus mean), but is kept for AI-IR as it is treated equal to human.
% \end{itemize}

\subsection{Model}
\label{sec:model}

\begin{figure}[h]
        \centering
        \includegraphics[width=1\textwidth]{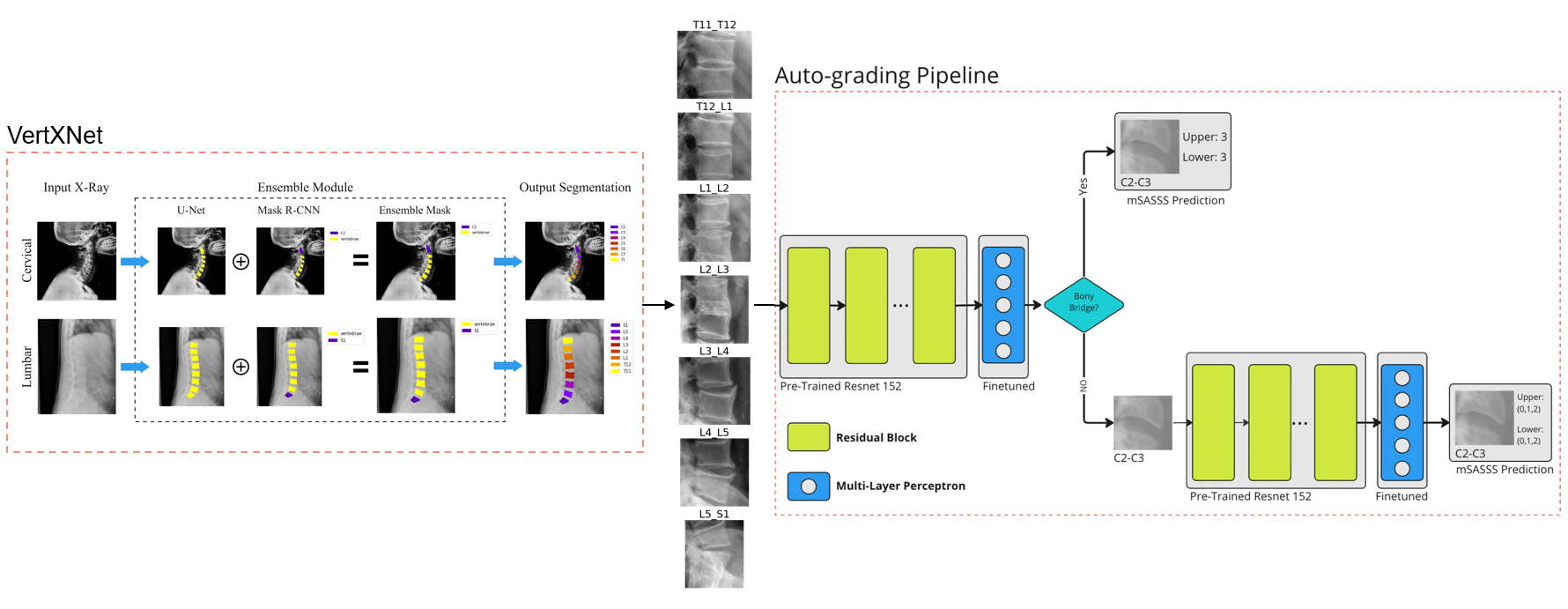}
        \caption{End to end pipeline to extract vertebral units (VUs) and mSASSS classification}
        \label{fig:pipeline}
    \end{figure}

% model introduction
A classification pipeline that incorporates vertebral unit segmentation \cite{chen2024vertxnet} followed by mSASSS classification \cite{mo2023towards} was developed as our AI reader (\Cref{fig:pipeline}). An ensemble model of nn-UNet \cite{ronneberger2015u, isensee2021nnu} and mask-RCNN \cite{he2017mask}, VertXNet\cite{chen2024vertxnet}, was used for vertebrae segmentation and labeling. All vertebral units were extracted on the basis of the segmentation results. Each vertebral unit would then be passed to a hierarchical grading system that consists of two ResNet 152 \cite{he2016deep} for the prediction of mSASSS.

% performance description
VertXNet has successfully extracted vertebral units with labels from 282 patients in MEASURE 1 and mSASSS were predicted by the auto-grading pipeline in \Cref{fig:pipeline}. mSASSS that could not be predicted from the pipeline were treated as missing as discussed in \Cref{sec:framework}. All mSASSS were predicted by models that were not trained with its original score. The grading model has achieved moderate performance of 65\% balanced accuracy which is the average of recall obtained on each class. 

\graphicspath{{./media/}}

\section{Results}

In this section, we compare the three frameworks mentioned in \Cref{fig:framework} from various perspectives, including cost, robustness, and study results. Our AI model, described in \Cref{sec:model}, serves as the basis for this comparison. To evaluate the robustness of these frameworks under less effective models, we also tested them using random and naïve models. The random model predicts individual mSASSS scores from 0 to 3 with equal probability, while the naïve model assigns a score of 0 to all mSASSS scores. These models simulate scenarios where a reasonable AI model is unavailable, allowing us to assess whether the frameworks can still produce robust results despite the bias introduced by the AI model.

\subsection{Cost}

In this section, we evaluated and compared the cost of using different frameworks that mentioned in \Cref{fig:framework}.
When relying solely on human readers, the results indicate that a second human reader and arbitration are required in 100\% and 48.92\% of cases, respectively, as shown in Table \ref{tab:cost}. For the AI model, it was 0.0\% and 59.83\% when AI is used as independent reader, and 59.83\% and 39.17\% when used as supportive reader. When comparing cost across the frameworks with the trained models from Section 2.3, both AI approaches (AI-IR, AI-SR) are more efficient than human alone (HDR) as they require fewer patient datasets to be read by a human reader..

    \begin{table}[h]
    \centering
    \begin{tabular}{|c|c|c|}
    \hline
    Method & Second human (\%) & Arbitration (\%) \\
    \hline
    HDR & 100 & 48.92 \\
    AI-IR & \textbf{0.0} & 59.83 \\
    AI-SR & 59.83 & \textbf{39.17} \\
    \hline
    \end{tabular}
    \caption{Percentage of patients requiring a second human reader and arbitration grading for the trained models.}
    \label{tab:cost}
    \end{table}

When stress testing our framework using a random model (each vertebral corner score is a random value between 0 and 3, uniformly distributed), we observed that AI-SR is matching the same cost as human alone - which is expected as safe-fail from the framework.

    \graphicspath{{./media/}}
    \begin{figure}[!htb]
    \centering
    \includegraphics[width=0.6\textwidth]{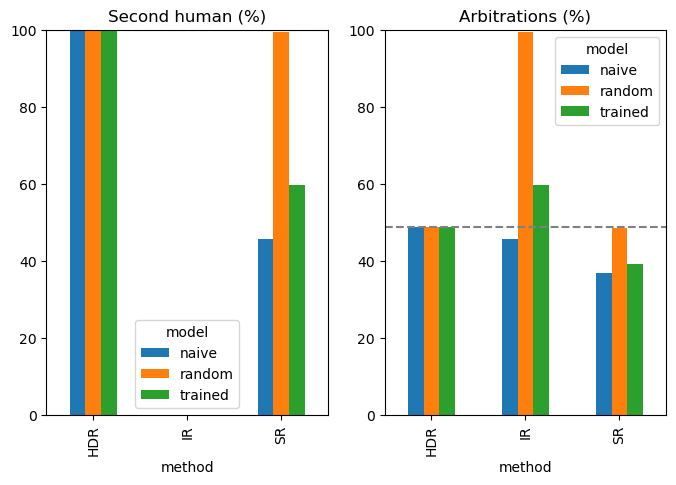}
    \caption{Cost difference between HDR, AI-IR and AI-SR when using extreme cases in prediction (naive and random model vs trained).}
    \label{fig:costs}
    \end{figure}

%Furthermore, we explored the hypothesis that arbitration cost is higher than the second human (e.g., need for more senior readers, multiple readers, etc..), ranging from 1 to 5 (folds) the cost of the second human reader. We found that in every case, both AI methods were outperforming the human. We also see that after 3 folds (e.g. cost of arbitration is three times higher than the second human reader), the AI-SR outperformed AI-IR and kept outperforming it as this ratio increases.

In practice, arbitration can be more costly than initial readings because it often requires the involvement of involvement of a an additional reader, increased experience levels or the initial readers to repeat their assessments. We investigated the scenarios that arbitration costs are significantly higher than those of a second human reader, with costs ranging from one to five times higher. Our findings revealed that both AI methods consistently outperformed human readers. Notably, when the arbitration cost reached three times that of the second human reader, AI-SR surpassed AI-IR and continued to maintain its superiority as the cost ratio increased.

    \graphicspath{{./media/}}
    \begin{figure}[!htb]
    \centering
    \includegraphics[width=1\textwidth]{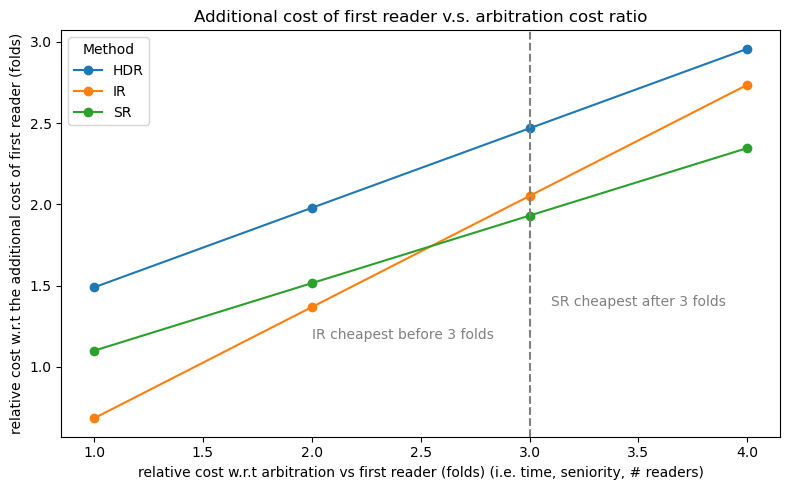}
    \caption{Impact of relative cost of arbitration w.r.t. the first human reader.}
    \label{fig:scaling}
    \end{figure}

\subsection{Model robustness}

\begin{figure}[!htb]
    \centering
    \includegraphics[width=0.6\textwidth]{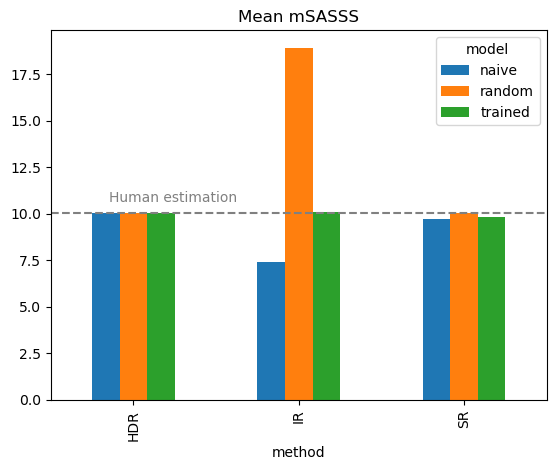}
    \caption{Mean mSASSS per patient accross all frameworks and models}
    \label{fig:mean}
    \end{figure}

The first and foremost quality of the framework is to be robust, which means that using AI should not change the disease evaluation. In this section, we compared the estimated mean mSASSS, across different models and methods in Figure \ref{fig:mean}. For the trained model, all methods yielded similar estimates (10.01, 10.09 \& 9.81) for the disease burden. But we found that AI-IR wasn't robust for extreme cases (i.e. random model), yielding 18.90, almost double the true estimate. This is because because the mSASSS consensus also uses the AI score, leading to a bias in the estimated value. For example, if you use a random model predicting either 0 or 1, the average value per sample is 0.5. Here for mSASSS, the average value is 1.5, because we randomly sample between 0 and 3 for each VU. On the other hand, when using a naive model (predicting only the majority class, for example here 0), the framework was much closer to the human, mostly due to the fact that the dataset predominantly includes patients with mSASSS vertebral corners of zero (approximately 80\%). 

    \graphicspath{{./media/}}

\subsection{Treatment effect}

In addition to the cost of the mSASSS reading and robustness of the frameworks, it is essential to verify that the conclusions drawn using these frameworks would align with those of human readers. Our simulation facilitated the comparison of results across real studies and different frameworks. In this section, we analyzed the clinical trial outcome, including summary statistics, endpoint distribution, and progression curves, across different frameworks, and compared them with the original trial results between treatment and placebo arms. 

The table of summary statistics and the conclusion of the hypothesis test on the endpoint are among the most crucial results of a trial. In our studies, worsening of mSASSS from baseline to 2 years was used as the endpoint to estimate the treatment effect. From \Cref{tab:stats}, it is evident that the conclusions on worsening are consistent across study outcomes and different frameworks. Although the control arm shows slightly worse worsening than the treatment arm, the difference is not statistically significant due to the large standard error. For other summary statistics, both HDR and AI-SR have demonstrated similar estimation with the trial results. However, AI-IR exhibits an overestimation bias at both baseline and week 104 mSASSS.

\begin{table}[h]
\centering
\begin{tabular}{|c|c|c|c|c|}
\hline
Method & Arm & Baseline & Week 104 & Worsening \\
\hline
\multirow{2}{*}{TRIAL} & Treatment & 10.79(17.48) & 10.98(17.49) & 0.54(4.73) \\
& Control & 9.49(16.26) & 11.87(18.20) & 0.91(4.45)\\
\hline
\multirow{2}{*}{Human double reader (HDR)} & Treatment & 10.85(18.09) & 
11.04(17.39) & 0.56(3.08)\\
& Control & 9.37(16.24) & 12.19(18.61) & 0.83(3.62)\\
\hline
\multirow{2}{*}{Independent reader (IR)} & Treatment & 11.15(17.84) & 11.26(17.38) & 0.49(2.95)\\
& Control & 9.64(16.14) & 12.36(18.47) & 0.77(3.68)\\
\hline
\multirow{2}{*}{Supportive reader (SR)} & Treatment & 10.75(18.13) & 10.99(17.40) & 0.51(3.11)\\
& Control & 9.36(16.25) & 12.11(18.64) & 0.78(3.63) \\
\hline
\end{tabular}
\caption{Trial results for MEASURE I. Human double reader has demonstrated very close reproduction of the trial results; Independent reader has over-estimation bias in both arms and time steps; Supportive reader has very close to the trial results on all estimation consider the large standard deviation.}
\label{tab:stats}
\end{table}

\begin{figure}[!htb]
       \begin{minipage}{0.48\textwidth}
         \centering
         \includegraphics[width=1\textwidth]{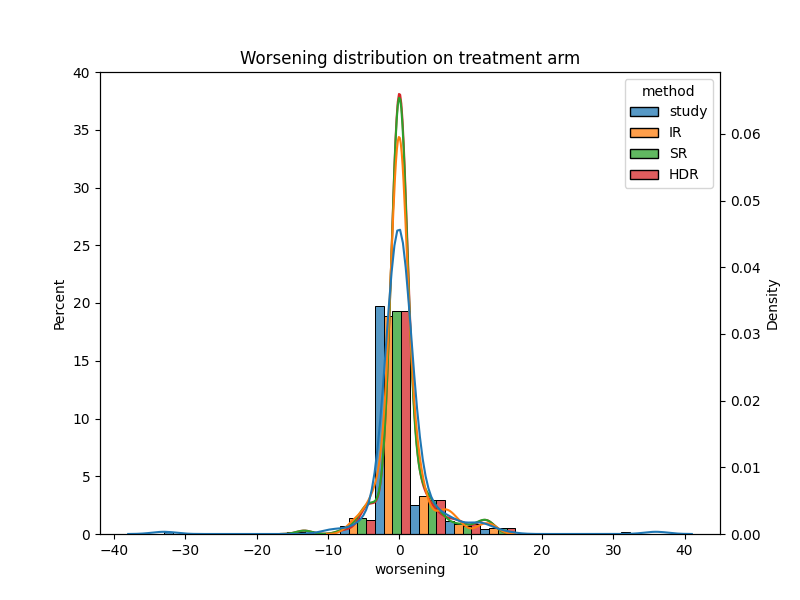}
         \caption{mSASSS worsening distribution in MEASURE I. mSASSS worsening distribution is plotted side by side for each bar; similar distributions of worsening were observed across all frameworks.}\label{fig:worsening_distribution}
       \end{minipage}\hfill
       \begin{minipage}{0.48\textwidth}
         \centering
         \includegraphics[width=1\textwidth]{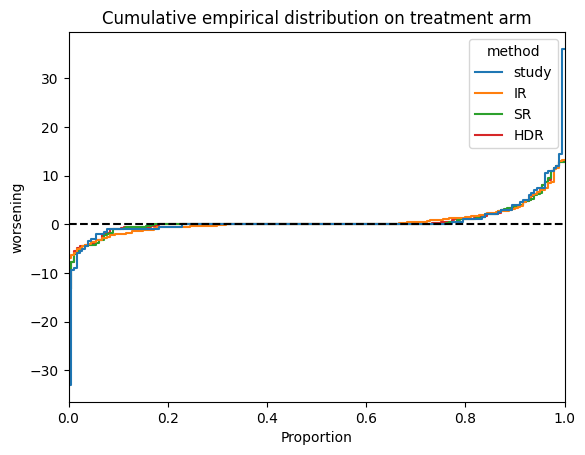}
         \caption{Empirical distribution of progression in MEASURE I. This figure shows the percentage of patients with worsening or improvement by a different extent displayed on the y-axis. All frameworks overlapping with each other pretty well.}\label{fig:progression_probability}
       \end{minipage}
    \end{figure}

We examined the distribution of mSASSS scores and the probability of progression in \Cref{fig:worsening_distribution} and \Cref{fig:progression_probability}. The overall distribution of mSASSS indicates that all frameworks are generating similar structural progression results as the trial results. \Cref{fig:progression_probability} depicts the proportion of patients experiencing worsening below the levels indicated on the y-axis. %A critical estimate, such as the percentage of patients with mSASSS progression, can be derived from this figure, where the overlapping curves demonstrate alignment with the conclusions.
From this figure, we can derive critical estimates, such as the percentage of patients with mSASSS progression, where the overlapping curves confirm alignment of the conclusions across all frameworks.

    \graphicspath{{./media/}}

All frameworks can produce the same statistical conclusions as the trial results regarding the hypothesis test. When AI acts as an independent reader, it has shown a systematic bias in its estimates on both arms. However, when AI serves as a supportive reader, it aligns quite well with the trial outcomes.

\subsection{Generalization}

\begin{figure}[!htb]
       \begin{minipage}{0.48\textwidth}
         \centering
         \includegraphics[width=1\textwidth]{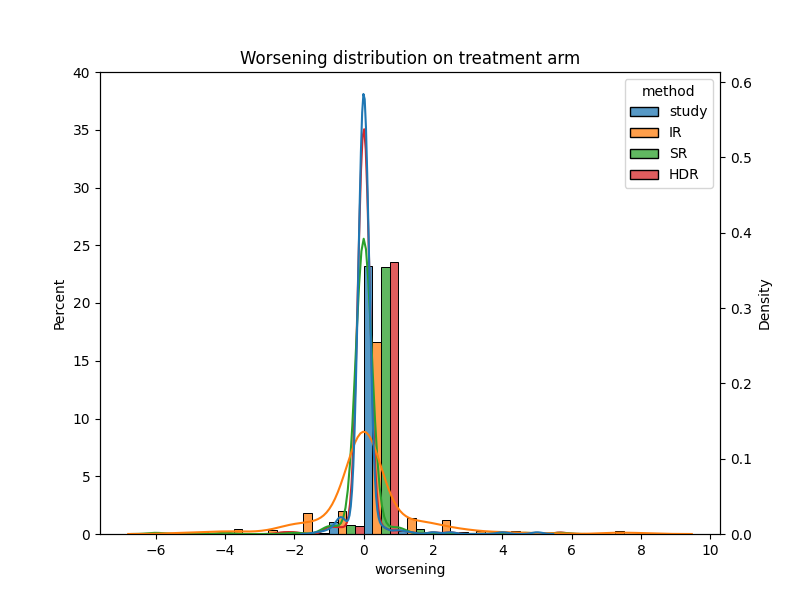}
         \caption{mSASSS worsening distribution in PREVENT. mSASSS worsening distribution is plotted side by side for each bar; similar distributions of worsening were observed across all frameworks.}\label{fig:worsening_distribution_prevent}
       \end{minipage}\hfill
       \begin{minipage}{0.48\textwidth}
         \centering
         \includegraphics[width=1\textwidth]{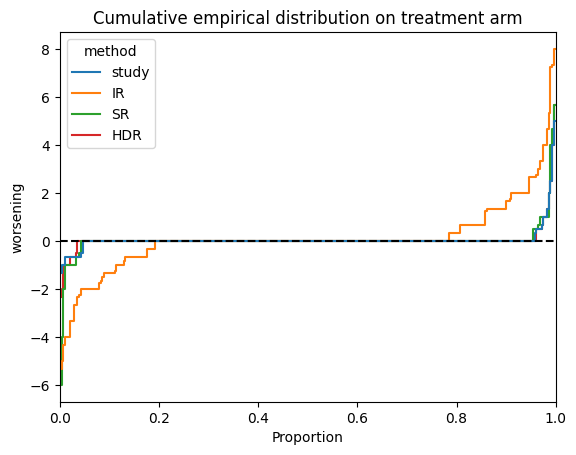}
         \caption{Empirical distribution of progression in PREVENT. This figure shows the percentage of patients with worsening or improvement by a different extent displayed on the y-axis. All frameworks overlapping with each other pretty well except AI-IR.}\label{fig:progression_probability_prevent}
       \end{minipage}
    \end{figure}

In reality, we seldom run two identical X-ray studies on design, disease area, doses or population. Even in the case of identical design and inclusion/exclusion criteria, the enrolled patient populations of the completed studies will be slightly different. To evaluate the potential benefits and robustness of AI-assisted frameworks in a new trial, we utilized the PREVENT trial. Given the differences of the patient population compared to MEASURE I and the more pronounced imbalance in patients' mSASSS readings, the AI grading model trained on MEASURE I would perform poorly on the PREVENT trial if used alone. However, our investigation revealed that all frameworks consistently reached the same conclusion as the trial results: there is no significant difference in worsening between the two arms. However, under close evaluation of the estimation and figures, AI-IR fails to generate convincing outcomes in this case. We displayed the same Table \ref{tab:stats_prevent} and Figure \ref{fig:worsening_distribution_prevent} / \ref{fig:progression_probability_prevent} on the PREVENT trial as follows.

\begin{table}[h]
\centering
\begin{tabular}{|c|c|c|c|c|}
\hline
Method & Arm & Baseline & Week 104 & Worsening \\
\hline
\multirow{2}{*}{TRIAL} & Treatment & 0.74(2.49) & 0.72(2.49) & 0.03(0.48) \\
& Control & 0.84(2.43) & 0.88(2.60) & 0.03(0.63)\\
\hline
\multirow{2}{*}{Human double reader (HDR)} & Treatment & 0.78(2.64) & 
0.76(2.65) & 0.03(0.53)\\
& Control & 0.84(2.49) & 0.97(2.86) & 0.09(0.46)\\
\hline
\multirow{2}{*}{Independent reader (AI-IR)} & Treatment & 1.18(2.64) & 1.26(2.75) & 0.09(1.51)\\
& Control & 1.31(2.68) & 1.29(3.00) & -0.10(1.41)\\
\hline
\multirow{2}{*}{Supportive reader (AI-SR)} & Treatment & 0.73(2.59) & 0.66(2.47) & 0.01(0.72)\\
& Control & 0.77(2.48) & 0.88(2.85) & 0.07(0.48) \\
\hline
\end{tabular}
\caption{Trial and model results for PREVENT. AI-IR performs poorly in the PREVENT population, leading to less accurate estimations. However, when used as a supportive reader, the AI provides estimations that are closely aligned with the trial results, despite the large standard deviation.}
\label{tab:stats_prevent}
\end{table}

 \graphicspath{{./media/}}

%The table and figures showed the generalizability of the framework that AI is used as supportive reader to other population in the same domain. AI-IR did not generalize well on the new trial with again over-estimation bias from \Cref{tab:stats_prevent} and different curve on progression probability in \Cref{fig:progression_probability_prevent}. AI-SR in general coincided with all perspectives of HDR as well as the trial results. It provided evidence that using an AI-assisted grading framework, specifically AI-SR, is valid in a new trial, even with a different population.

The table and figures demonstrate the generalization ability of the framework where AI is used as a supportive reader (AI-SR) for other populations in the same domain. In contrast, AI-IR did not generalize well to the new trial, showing an overestimation bias in the table and a differing progression probability curve in \Cref{fig:progression_probability_prevent}. On the other hand, AI-SR closely aligned with all perspectives of the HDR and the trial results. This provides evidence that the AI-assisted grading framework, particularly AI-SR, is valid and reliable in a new trial, even with a different population.

\section{Discussion}

\begin{table}[]
    \centering
    \begin{tabular}{c|c|c|c}
         Framework & HDR & AI-IR & AI-SR \\
         \hline
         Cost & \xmark & \cmark & \cmark \\
         Robustness & \cmark & \xmark & \cmark \\
         Generalization & \cmark & \xmark & \cmark \\
         Accuracy & \cmark & \pmark & \cmark \\
        \hline
    \end{tabular}
    \caption{Overall evaluation of all methods. AI as supporting reader (AI-SR) has proven to be the most robust, efficient and accurate method with all models. \cmark $\,$ indicates success, \xmark $\,$ indicates failure, \pmark $\,$  indicates a partial success.}
    \label{tab:summary}
\end{table}

In this study, we investigated the potential application of AI in clinical trials. We characterized that an effective AI framework should meet three key criteria: accuracy, efficiency, and robustness. First, the AI framework should produce the same trial results as those obtained by human-only graders. Second, the AI framework should save time and money compared to a human-only grading system. Lastly, the AI framework should deliver consistent evaluations even when its model performance is suboptimal. To test these criteria, we experimented with three different methods under varying model performances: trained, random, and naive. This approach allowed us to stress test the limits of each method within the framework.

Our research found that the AI as supportive reader (AI-SR) approach is the most suitable for clinical trials, as it meets all criteria across all model types. This framework excels in accuracy because it places AI under human supervision, ensuring safety and efficiency by reducing the number of human readers needed. It also provides accurate estimations of disease status and treatment effects. 
One finding is that we don't need a high performing model to see any gain in the efficiency while keeping accurate results. Like ensemble method, it does not require a strong individual grader, AI or human, but still performs consistently on the final output. In practice, when the sample size is small or the outcome is imbalanced, only moderately performing models are achievable. In such a case, it is recommended that AI serves as a supportive reader that has less weight on the final grading. This framework would still reduce trial cost even when the arbitrator is much more expensive than the first/second reader. At the same time, AI graders could also be updated by pooling more studies. As their performance enhances, the weight of the AI graders could also be elevated. Consequently, AI serving as independent reader could be more reliable.

%To the best of our knowledge, this is the first study observing the effect of AI assisted grading framework on the evaluation of treatment effect and trail conclusion from a clinical trial. This result is the most important as it compound the total effect of AI in the trial.

To the best of our knowledge, this is the first study to examine the impact of an AI-assisted grading framework on the evaluation of treatment effects and trial conclusions in a clinical setting. This finding is particularly significant as it highlights the importance of the choice of the right AI framework for assessing outcomes of clinical trials.

% \section{FUTURE STEP}

% Future works will aim to confirm the findings in other clinical trials, such as SURPASS 

% USE SURPASS as next step, future works

%Bibliography
\bibliographystyle{unsrt}  
\bibliography{references}

\end{document}